\documentclass{article}

    \usepackage[preprint]{neurips_2024}

\usepackage[utf8]{inputenc} 
\usepackage[T1]{fontenc}    
\usepackage{hyperref}       
\usepackage{url}            
\usepackage{booktabs}       
\usepackage{amsfonts}       
\usepackage{nicefrac}       
\usepackage{microtype}      
\usepackage{xcolor}         
\usepackage{graphicx}
\usepackage{amsmath}
\usepackage{cleveref}
\usepackage{subcaption}

\title{Are PPO-ed Language Models Hackable?}

\usepackage{authblk}

\makeatletter
\renewcommand\AB@affilsepx{  \protect\Affilfont}

\makeatother

\author[ ]{Suraj Anand}
\author[ ]{David Getzen}
\affil[ ]{\authorcr \protect\Affilfont Brown University}
\affil[ ]{\authorcr \protect\Affilfont Department of Computer Science}
\affil[ ]{\authorcr \protect\Affilfont Correspondence to \texttt{suraj\_anand@brown.edu}}

\begin{document}

\maketitle

\begin{abstract}
  Numerous algorithms have been proposed to \textit{align} language models to remove undesirable behaviors. However, the challenges associated with a very large state space and creating a proper reward function often result in various jailbreaks. Our paper aims to examine this effect of reward in the controlled setting of positive sentiment language generation. Instead of online training of a reward model based on human feedback, we employ a statically learned sentiment classifier. We also consider a setting where our model's weights and activations are exposed to an end-user after training. We examine a pretrained GPT-2 through the lens of mechanistic interpretability before and after proximal policy optimization (PPO) has been applied to promote positive sentiment responses. Using these insights, we (1) attempt to "hack" the PPO-ed model to generate negative sentiment responses and (2) add a term to the reward function to try and alter `negative' weights.
\end{abstract}

\section{Introduction}

Large language models have powerful emergent capabilities from self-supervised training on natural datasets \citep{brown2020, wei2022emergent}. This training results in high performance next-word prediction; however, as most are trained on large internet corpuses \citep{gao2020pile, bookscorpus2015}, vanilla training results in a number of undesirable behaviors such as toxicity, bias, negative sentiment, and instruction-following failure \citep{sheng-etal-2019-woman, gehman-etal-2020-realtoxicityprompts}. These shortcomings have led to the use of post-training patching to correct undesirable behaviors. Namely, various algorithms have been proposed to \textit{align} language models under the hood of reinforcement learning from human feedback (RLHF) \citep{kaufmann2024survey}.

RLHF introduces a critical online, human-in-the-loop component to the standard language model learning paradigm. In standard reinforcement learning, an agent attempts to make optimal decisions based off a reward function that it aims to maximize \cite{sutton2018reinforcement}. However, specifying this reward function is often quite challenging; sparse rewards result in the need for \textit{reward shaping} and loopholes result in the issue of \textit{reward hacking} \citep{Ng1999PolicyIU, skalse2022hacking}. 

The challenges associated with creating a proper reward function often result in various jailbreaks \citep{rao2024tricking, li2023multistep}. Our paper aims to examine this effect of reward in the controlled setting of positive sentiment language generation. Instead of online training of a reward model based on human feedback, we employ a statically learned sentiment classifier. We also consider a setting where our model's weights and activations are exposed to an end-user after training. 

We rely on recent insights from the field of mechanistic interpretability to jailbreak an aligned model \citep{olah2022mechanistic}. This field focuses on reverse-engineering model components into human understandable algorithms, and usually involves neuron, weight, and activation level analysis to do so. In this discipline, we view a model as a computational graph based on soft cause-and-effect relationships \citep{geiger2021causal}. In this research, we examine a pretrained GPT-2 through this perspective before and after reinforcement learning has been applied. From this, we glean a causal understanding of model changes and insight into how to potentially "hack" an aligned GPT-2. Moreover, we attempt to utilize these insights to modify our reward function by adding a loss term that targets certain weights in our model.

Our main contributions are:

\begin{enumerate}
    \item We analyze what causal changes PPO results in on GPT-2 at an activation and weight level.
    \item We attempt to engineer a methodology that circumvents these changes to output negative sentiment sequences. 
    \item We experiment with the reward function to try and mitigate this jailbreak.
\end{enumerate}

\begin{figure}[t!]
    \centering
    \vspace{1cm}
    \fbox{
    \includegraphics[width=0.8\linewidth]{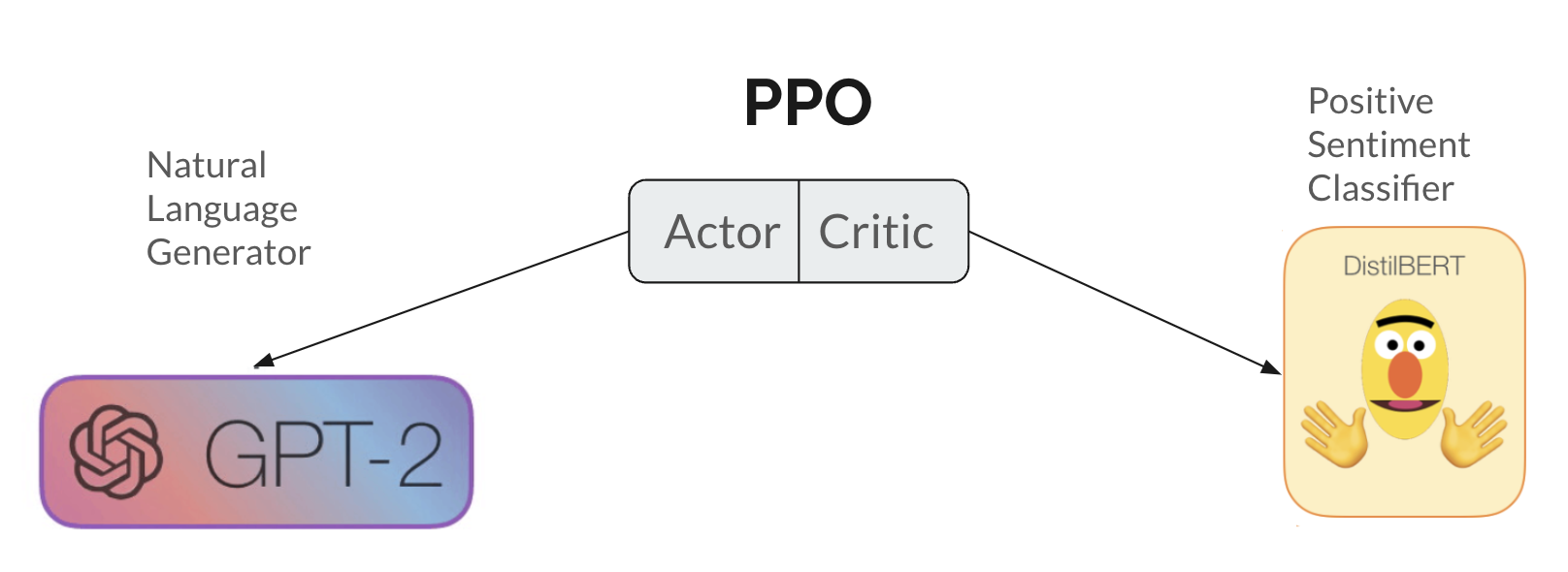}
    }
    \caption{Our controlled setup employs a pretrained GPT-2 model to generate sequences (rollouts) and a DistilBERT classifier to evaluate the sentiment (reward). }
    \label{fig:ppo_actor_critic}
\end{figure}

\section{Related Works}
Our contributions build upon recent and concurrent research in the field. \citep{gehman-etal-2020-realtoxicityprompts} establishes the urgency of toxicity avoidance as a problem space, showcasing the ease with which pre-trained language models generate outputs reflecting toxic sentiment given prompts of any kind. 

A framework for leveraging word vectors to capture sentiment is presented in \citep{maas-etal-2011-learning}, wherein sentiment polarity annotations, such as metrics that correspond to reviews or critiques, are utilized in the generation of sentiment predictions with significant success. This project additionally introduces a sizable dataset of movie reviews extracted from the service IMDb to be used within the context of sentiment classification. 

We are motivated by the mechanistic unalignment methodology found in \citep{lee2024mechanistic}. This research discovered that a model aligned with Direct Preference Optimization to avoiding toxicity in its outputs accomplishes favorable returns only by learning an offset. As such, the model avoids internal representations of vectors that would trigger output toxicity rather but does not remove toxic weights altogether. We extend this research to (1) consider positive sentiment generation, (2) examine PPO, and (3) introduce a loss function to attempt to remedy this shortcoming.

We build off internal representation translation work that began with \cite{nostalgebraist_interpreting_gpt}, which introduced a "Logit Lens", or an un-embedding mechanism that allows the projection of token representations from an intermediate embedding space to a final vocabulary space, enabling the observation of tokens that are generated in each subsequent embedding layer within a transformer-based language model. A closer glimpse at the internal behavior of a transformer-based language model is found in \citep{geva-etal-2022-transformer} where it was observed that tokens are hierarchically ordered through each sub-update in the feed-forward embedding layers of transformer-based language models, with each intermediary output token being assigned a certain value-vector $v_{\mathbf{i}}$ corresponding to the probability of its expression in the next embedding layer. 

Within the context of our project, we aimed to leverage the method proposed by \citet{geva-etal-2022-transformer} as a means of finding weights with negative sentiment, then inflating their corresponding activations to promote their associated concepts within the final output of a pretrained, transformer-based language model, un-aligning the impact of PPO on fitting it to the task of output toxicity.  

\section{Preliminaries}

We borrow much of the transformer and vocabulary space notation from \cite{lee2024mechanistic} and \cite{geva-etal-2022-transformer} to express our experiments. 

\subsection{Transformers}

We employ an encoder-only transformer-based language model. Transformers consist of an embedding layer, $E \in \mathbb{R}^{|V| \times d}$, a series of $L$ transformer layers, and then an unembedding layer $U \in \mathbb{R}^{|V| \times d}$ which can be tied to the embedding layer \citep{vaswani2023attention}. Each layer $\ell$ consists of multiple attention heads as well as a multilayer perceptron (MLP) layer.

Given a sequence of input $w = \{w_0, \ldots, w_t\}$, the transformer first applies $E$ to create an embedding $x_i \in \mathbb{R}^d$ for each token $w_i \in w$.

This representation is then updated by attention heads and MLP blocks each layer that follows:
\begin{equation*}
    x_{i+1}^\ell = x_i^\ell + \mathrm{MLP}^\ell(x_i^\ell + \mathrm{Att}^\ell(x_i^\ell))
\end{equation*}
We denote this token-wise representation as the "residual stream," following previous literature. In our encoder-only GPT structure, the updates to the residual stream from each MLP block are able to be further decomposed into two linear transformations with nonlinear activations in between:
\begin{equation*}
    \mathrm{MLP}^\ell(x^e) = \sigma\left(\mathbf{W}_K^\ell x^e\right) \mathbf{W}_V^\ell
\end{equation*}
where $\mathbf{W}_K^\ell, \mathbf{W}_V^\ell \in \mathbb{R}^{d_{\mathrm{mlp}} \times d}$. We call the $i$-th row in $\mathbf{W}_K^\ell$ as $\mathrm{MLP}_i^\ell x^e$ and refer to them as key-vectors, and the $i$-th column in $\mathbf{W}_V^\ell$ as value-vectors (we sometimes omit “MLP” and just use $k_i^\ell, v_i^\ell$). The output of MLP blocks is the sum of its value vectors $v_i$, each scaled by a coefficient value $m_i^\ell$, where $m_i^\ell = \sigma\left(\mathbf{W}_K^\ell x^e\right)_i$.

\subsection{Vocabulary Space Translation}
\label{sec:vocab_space_prelim}
\cite{geva-etal-2022-transformer} demonstrate that as the residual stream is processed, each value vector $v_i$ either promotes or suppresses the likelihood of a token $w$ to be generated:
\[p\left(w \mid x^e + m_i^\ell v_i^\ell, E\right) \propto \exp\left(e_w \cdot x^e\right) \cdot \exp\left(e_w \cdot m_i^\ell v_i^\ell\right)\]
where $e_w$ is the embedding of $w$. As such, when $e_w \cdot m_i^\ell v_i^\ell > 0$ we see that the likelihood of $w$ increases, while $e_w \cdot m_i^\ell v_i^\ell < 0$ decreases the likelihood of $w$.\footnote{Note that $e_w \cdot v_i^\ell$ does not depend on the input and it is only when $v_i^\ell$ is scaled by $m_i^\ell$ result in any change to output distribution.}

\subsection{Reward Function for PPO}
\label{sec:ppo}
The objective of Proximal Policy Optimization (PPO) is to maximize a reward function \(r(\theta)\) using a policy \(\pi_\theta\) parameterized by \(\theta\) \citep{schulman2017proximal}. Given a static reward model \(R(w)\) (our sentiment classifier) on GPT-2, where \(w\) represents a sequence of tokens generated by the model, the reward function can be defined as follows:

\begin{equation*}
    r(\theta) = \mathbb{E}_{w \sim \pi_\theta}[R(w)]
\end{equation*}

PPO seeks to maximize this reward while keeping the policy changes between updates constrained, to avoid large deviations from the existing policy. The probability of generating a sequence \(w\) under policy \(\pi_\theta\) is denoted as \(\pi_\theta(w)\). 

\subsubsection{PPO Objective}

The PPO objective \(L(\theta)\) can be expressed as:

\begin{equation*}
    L(\theta) = \mathbb{E}_{w \sim \pi_{\theta_{\mathrm{old}}}}\left[\min\left(\frac{\pi_\theta(w)}{\pi_{\theta_{\mathrm{old}}}(w)}A(w), \mathrm{clip}\left(\frac{\pi_\theta(w)}{\pi_{\theta_{\mathrm{old}}}(w)}, 1 - \epsilon, 1 + \epsilon\right)A(w)\right)\right]
\end{equation*}

where:
\begin{itemize}
    \item \(A(w)\) is an advantage function that estimates how much better \(w\) is compared to other possible sequences.
    \item \(\epsilon\) is a hyperparameter controlling the clipping range.
\end{itemize}

The above objective ensures that policy updates are made conservatively to prevent large changes, which aligns the PPO policy to the static reward model without major changes to output distribution. The rewards as rollout is calculated are KL penalties and the final reward after the EOS token adds the sample score. This is a simpler and more efficient way to limit policy changes compared to TRPO \citep{schulman2017trust}.

\subsection{Combined PPO Reward Function}

To combine the reward function \(r(\theta)\) with the PPO objective, the final reward function can be expressed as:

\begin{equation*}
    r_{\text{ppo}}(\theta) = L(\theta) + \lambda_1 \mathbb{E}_{w \sim \pi_\theta}[R(w)]
\end{equation*}

where \(\lambda_1\) is a hyperparameter balancing the two components of the reward function.

\begin{figure}[h!]
    \centering
    \vspace{0.2cm}
    \fbox{
    \includegraphics[width=0.8\linewidth]{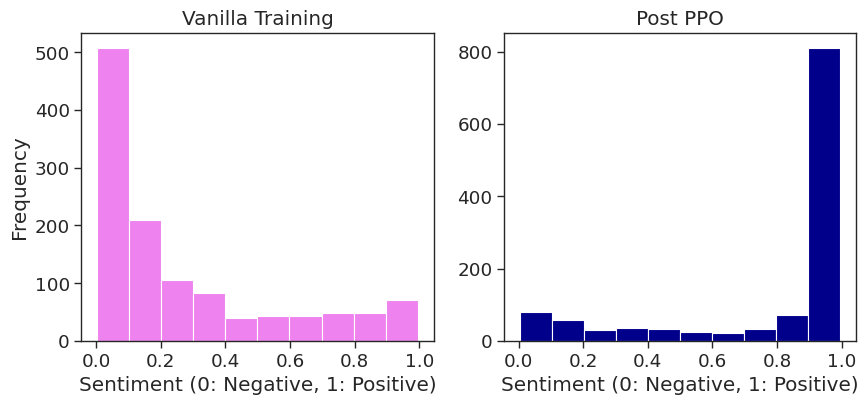}
    }
    \caption{Histogram of sentiment of responses in heldout test set pre and post PPO the GPT-2. }
    \label{fig:sentiment_change} 
\end{figure}

\section{Sentiment in GPT-2}

Throughout our experimental process, we used a GPT2 model fine-tuned on the Stanford Natural Language Processing Large Movie Review Dataset, which consisted of 50,000 highly polar film reviews extracted from the service IMDb. The dataset assessed the sentiment of a review as either "positive" or "negative".  

\subsection{Negative Weights}

With \cref{sec:vocab_space_prelim} as a preface, we assessed negativity in weights by examining the value-vectors corresponding to each token. 

To do so, we trained a linear probe to discern negativity on the last-layer internal representations of sentences (created by token-wise averaging) from the IMDb sentiment dataset \citep{imdb2011dataset}. We extracted the vector $W_{\text{neg}}$ from the probe weight. This is the vector orthogonal to the hyperplane that separates negative and positive sentiment representations. To find the value-vectors $v_l^i$ affiliated with the most negative tokens consisted of finding those with the highest cosine similarity to $W_{\text{neg}}$ generated from the $k$ most negative value weights within the leveraged dataset. 

Consulting \cref{tab:vocab_space}, it can be discerned that a group of the most negative weights, or those with the highest cosine similarity with our negative probe vector, correspond to sets of tokens comprised of phrases that reflect significantly negative sentiment. Each token-set has a negatively-ranked value-vector assignment, and this particular group of the most negative token-sets are positioned across layers 6 through 9 of our pre-trained, un-aligned GP2 model. Interestingly, we found that the vast majority of `negative' value vectors were located in the second half of the layers in the model.

\begin{table}[h!]
  \caption{`Negative' Weights in GPT-2 and the PPO-ed GPT-2 mapped into the vocabulary space by using the method proposed by \cite{geva-etal-2022-transformer}}
  \label{tab:vocab_space}
  \vspace{10pt}
  \centering
  \begin{tabular}{llll}
    \toprule
    \textbf{Layer}     & \textbf{Index} & \textbf{GPT-2} & \textbf{PPO-ed GPT-2} \\
    \midrule
    7 & 2394 & useless, mediocre, worthless & useless, mediocre, worthless \\
    6 & 2360 & unus, disastrous, deteriorated & unus, disastrous, deteriorated \\
    9 & 3047 & negligible, diminished, fewer & negligible, diminished, fewer \\
    7 & 2464 & fail, Fail, Wrong & fail, Fail, Wrong \\
    8 & 2635 & bad, horrendous, problematic & bad, horrendous, problematic \\
    9 & 484 & insufficientlessness, inability, incorrect & insufficientlessness, inability, incorrect \\
    \bottomrule
  \end{tabular}
\end{table}

\section{Promoting Positivity with PPO}

We perform PPO (described in \cref{sec:ppo}) in order to promote positive sentiment responses by GPT-2. In our formulation, we employ a DistilBERT \citep{sanh2020distilbert} finetuned on the IMDb sentiment dataset to classify sentiment as our reward model. We employ the \texttt{trlx} library to perform this reinforcement learning \citep{havrilla-etal-2023-trlx}. We find that the average sentiment score across a heldout prompt set was 0.27 for the original model and 0.80 on the PPO-ed model. This shows that our model can now successfully generate positive sentiment responses to a diverse set of input prompts.

To get a better understanding of what changed under the hood, we perform a mechanistic analysis of the pre and post PPO-ed model. This helps us causally determine how the model was changed, and whether its alignment with the task of output toxicity occured only due to learning an offset that would enable the avoidance of toxicity-activating vectors. With a finer glimpse into how positivity is promoted by a PPO-finetuned GPT-2, a foundation for efforts toward preventing its mechanistic un-alignment with respect to a given task can be formed.  

\begin{figure}[h!]
    \centering
    \vspace{0.2cm}
    \fbox{
    \includegraphics[width=0.8\linewidth]{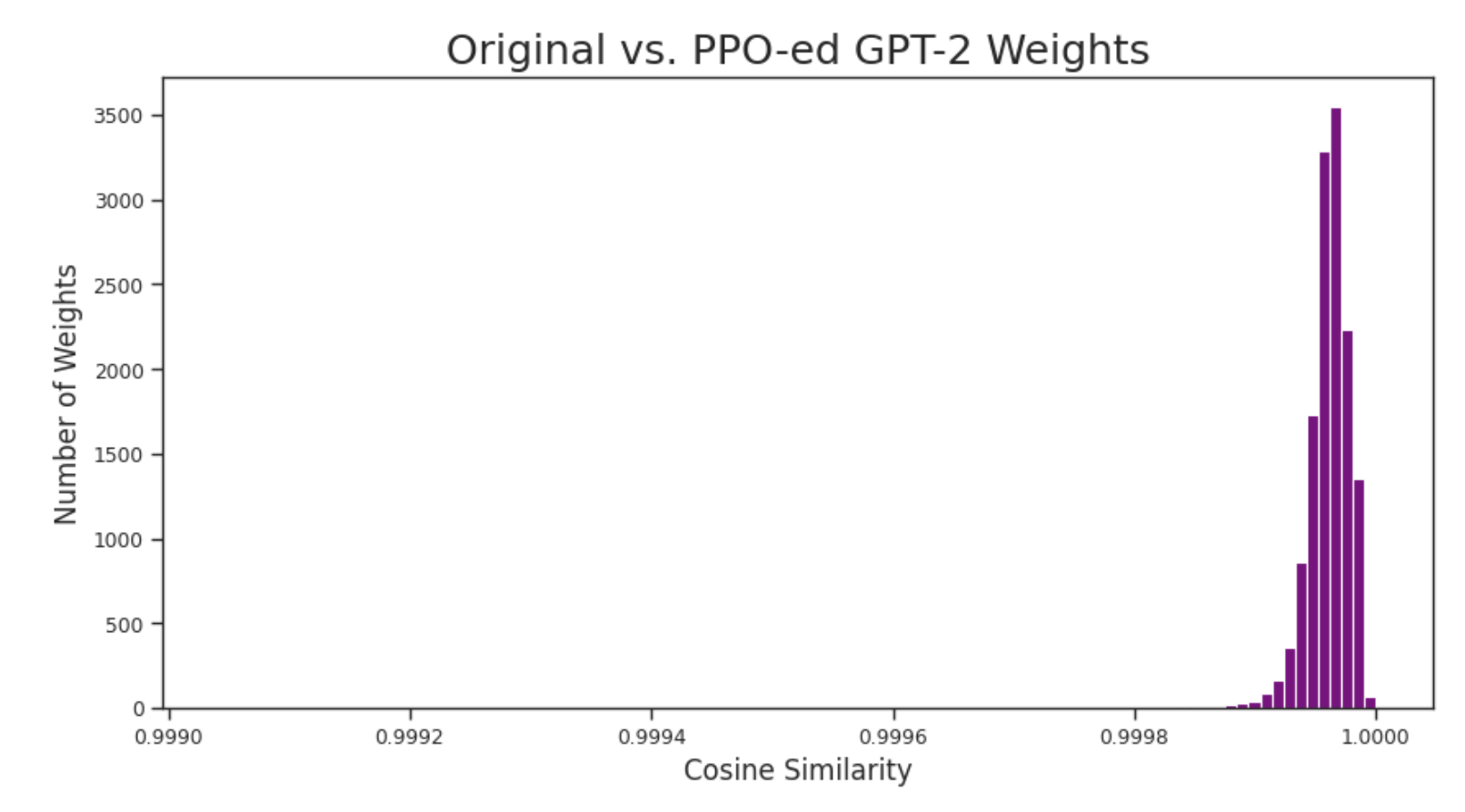}
    }
    \caption{Weights are minimally changed by Proximal Policy Optimization of the full GPT-2 model. }
    \label{fig:weight_change} 
\end{figure}

\subsection{Weights are Minimally Altered}

As evident by \cref{fig:weight_change}, the weights of GPT-2 are only very slightly modified by PPO. In fact, almost all modified weights remain at a cosine similarity of $\ge 0.9998$ with their original values. It was surprising to us that this little of a change to weights can cumulatively result in a change in behavior toward positive sentiment results. Additionally, we ran experiments where only a subset of layers were allowed to change with PPO (2 layers, 4 layers). In these experiments, the cosine similarity values were slightly lower, but still well over 0.9995 for the significant majority.

\subsection{Negative Weights Remain}

We find that the `negative' weights discovered earlier in our research remain almost completely unchanged in \cref{tab:vocab_space}. This suggests that PPO learns a wrapper around the model that is similar to the offset that is found in \cite{lee2024mechanistic}. This is suboptimal as it means that negative concepts may still be stored within the model's weights which are not removed from PPO. However, it is expected as PPO used a clipped objective that tries to make only very small changes to the policy.

Moreover, this suggests that we might be able to still promote concepts in the negative sentiment space by simplying scaling up the coefficients $m_i^l$ that correspond to each negative value vector (or similarly by increasing the activation after this vector). We test this intervention in the following section.

\begin{figure}[h!]
    \centering
    \fbox{
      \begin{subfigure}[b]{0.52\textwidth}
        \centering
        \includegraphics[width=\textwidth]{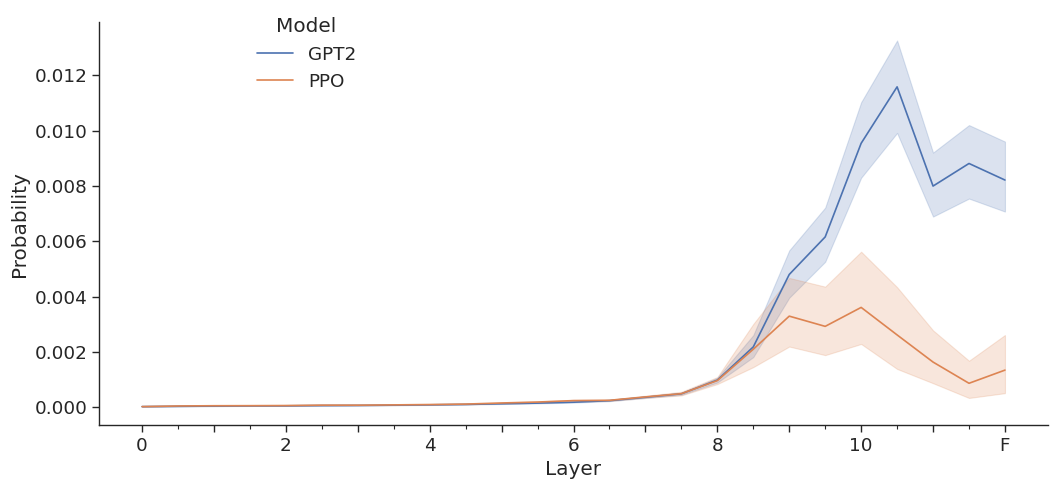}
        \label{fig:shit_pred}
    \end{subfigure}
    \hfill
    \begin{subfigure}[b]{0.47\textwidth}
        \centering
        \includegraphics[width=\textwidth]{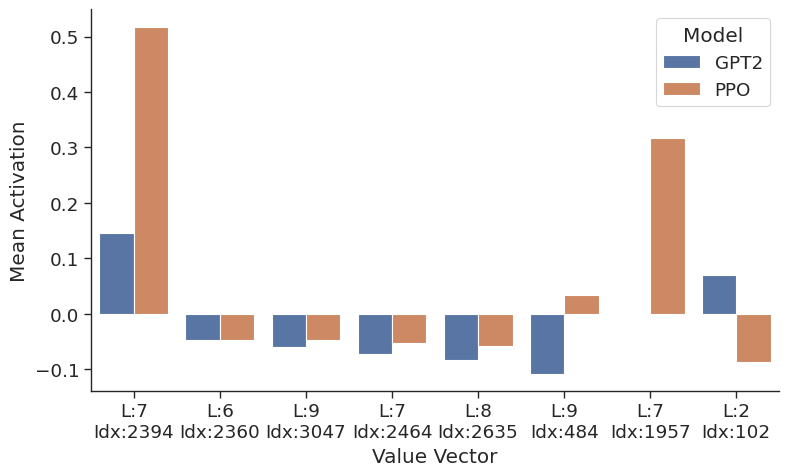}
        \label{fig:act_change}
    \end{subfigure}
    }
    
    \caption{(Left) Logit lens shows that negative concept activations scaled down across model residual stream (this is specifically for the word sh*t). (Right) The activation differences for the ten most `negative' value vectors. }        
    \vspace{-10pt}
    \label{fig:activation_fig} 
\end{figure}

\subsection{Activations Change}
\label{sec:activations}

We utilize the logit lens method to determine the extent of negative concept activation change over time \citep{nostalgebraist_interpreting_gpt}. Essentially, to do this, we take the softmax of $Ux^l_{i+1}$ (i.e. the unembedding matrix multiplied by the residual stream of the next token at layer $i$). We find a common trend that concepts overall were less promoted in the residual stream after PPO. This is consistent with the fact that our PPO successfully results in positive sentiment changes. Interestingly, we find that this change in concept promotion happens throughout the model when all layers are fine-tuned, rather than simply in the last couple of layers.

In addition, we quantify how much the activations change from before and after PPO for a heldout set of the negative response generating prompts for the top $k$ `negative' value vectors. To do this, we simply take the delta of the two values. The results can be seen in \cref{fig:act_change}. Surprisingly, the activations are not consistently lower magnitude, but instead some `negative' weights are activated to a higher degree. We hypothesize that this might be due to the fact that there is some randomness in the weight selection; moreover, we still find that overall `negative' concepts are promoted less in the residual stream. Thus, even though some activations might be higher, in total they result in less strength in the residual stream.

\subsection{Activation Scaling}
To "hack" PPO, we try scaling the key values that correspond to each of the top $k$ negative value vectors by a factor of 10  to promote negative concepts in the unembeddings space to be added to the residual stream. We were slightly unsure whether this would work due to our strange results in \cref{sec:activations}. What we found was that this did effectively promote negative concepts and revert the positive changes of PPO. The mean sentiment on our heldout test set went from 0.80 to 0.43, which signifies that we effectively jailbroke the model to generate negative sentiment outputs. Note that this required access to the weights a priori. 

\subsection{Regularization}

To mitigate the issue that our model simply wraps computation to avoid stored `negative' weights rather than removing `negative' weights, we add a term to the reward function that rewards this behavior. In a simple depiction, this can be thought of as

\begin{equation*}
    r = r(\mathbf{\Theta}) - \lambda_1 r_{KL}
\end{equation*}

\begin{equation*}
    \downarrow
\end{equation*}

\begin{equation*}
    r = r(\mathbf{\Theta}) - \lambda_1 r_{KL} + \lambda_2 \sum_{w \in \mathbf{N}} \|w - w_{\text{original}}\|
\end{equation*}

where $w \in \mathbf{N}$ are the negative weights in the negative weight set, where this set has a cardinality of 10. We altered trlx code in our to get this functionality possible. We chose a $\lambda_2 = 10^{-4}$, but played around with it a little bit. We added this reward to the end of the rollout (at the same time the positive sentiment reward is calculated). We were able to achieve results that altered the weights, but were not able to find a stable hyperparameter that was a good medium between (1) altering the weights enough to meaningfully store remove the negative information and (2) maintaining the stability of the output distribution. When the parameter for removing negative weights was too high, it would cause instability in the output generation and break natural language function. Overall, we think that this is a good approach, but need to better conduct a more rigorous and fine-grained search over the hyperparameter space of $\lambda_2$

\section{Discussion}

In this work, we studied a mechanistic account of how reinforcement learning alignment algorithms alter models. We specifically took PPO, and its use as an alignment algorithm in fitting a pre-trained GPT2 to the task of avoiding negative sentiment, as a case study. To do so, we uncovered how negative sentiment is represented and elicited in a pre-trained language model. Applying the insights provided into the use of value-vectors within the context of transformer-based language models by \citep{geva-etal-2022-transformer}, we found numerous vectors in MLP blocks that promote negative sentiment, for which subtracting these vectors from the residual stream can suppress negative outputs. We then applied PPO to our language model to alter model weights to promote positive sentiment outputs successfully. Finally, we studied how our aligned GPT-2 averts negative sentiment. Rather than removing `negative' weights, the PPO-ed GPT-2 alters the activations of a number of very, very small changes to the weights, but in a manner that is somewhat surprising to us. That is, the activations of a few negative weights were somewhat higher after the application of PPO. However, with this as an exception, weights that corresponded to toxicity were, on average, promoted less throughout the progression of the residual stream. While the specific behavior of the PPO-ed GPT2 differed somewhat from the DPO-centered findings of \citep{lee2024mechanistic}, the overall potential for un-alignment, or "hackability", was demonstrated nonetheless in our findings due to a similar tendency of PPO to facilitate only an avoidance of toxic activation regions in a transformer-based language model. 

This enables the model to preserve its pretrained behavior while averting negative outputs. This is expected as PPO results in small policy changes that minimally change the output distribution.

Given this understanding, we show how to "hack" the alignment of the PPO-ed GPT-2. That is, we simply increase the regions that elicit negative sentiment by scaling their corresponding key vectors. We find this hack successful. Finally, we attempt to alter the reward signal with mechanistic information to remove negative stored weights from our model. This methodology showed promise, but a more intensive hyperparameter search is necessary in order to balance stability of natural language output distribution generation and removing negative weights. 

Overall, this was a mechanistic case study for how Proximal Policy Optimization alters models, whether we can jailbreak these models, and one simple attempt to patch this. We hope that in the future this can
act as a stepping stone to more robust alignment algorithms.

\section*{Acknowledgements}
We conducted this research as the final project for CSCI2951F: Learning and Sequential Decision Making at Brown University. This research was conducted using computational resources and services at the Center for Computation and Visualization, Brown University.

\bibliography{references}

\begin{thebibliography}{24}
\providecommand{\natexlab}[1]{#1}
\providecommand{\url}[1]{\texttt{#1}}
\expandafter\ifx\csname urlstyle\endcsname\relax
  \providecommand{\doi}[1]{doi: #1}\else
  \providecommand{\doi}{doi: \begingroup \urlstyle{rm}\Url}\fi

\bibitem[Brown et~al.(2020)Brown, Mann, Ryder, Subbiah, Kaplan, Dhariwal, Neelakantan, Shyam, Sastry, Askell, Agarwal, Herbert{-}Voss, Krueger, Henighan, Child, Ramesh, Ziegler, Wu, Winter, Hesse, Chen, Sigler, Litwin, Gray, Chess, Clark, Berner, McCandlish, Radford, Sutskever, and Amodei]{brown2020}
Tom~B. Brown, Benjamin Mann, Nick Ryder, Melanie Subbiah, Jared Kaplan, Prafulla Dhariwal, Arvind Neelakantan, Pranav Shyam, Girish Sastry, Amanda Askell, Sandhini Agarwal, Ariel Herbert{-}Voss, Gretchen Krueger, Tom Henighan, Rewon Child, Aditya Ramesh, Daniel~M. Ziegler, Jeffrey Wu, Clemens Winter, Christopher Hesse, Mark Chen, Eric Sigler, Mateusz Litwin, Scott Gray, Benjamin Chess, Jack Clark, Christopher Berner, Sam McCandlish, Alec Radford, Ilya Sutskever, and Dario Amodei.
\newblock Language models are few-shot learners.
\newblock \emph{CoRR}, abs/2005.14165, 2020.
\newblock URL \url{https://arxiv.org/abs/2005.14165}.

\bibitem[Gao et~al.(2020)Gao, Biderman, Black, Golding, Hoppe, Foster, Phang, He, Thite, Nabeshima, Presser, and Leahy]{gao2020pile}
Leo Gao, Stella Biderman, Sid Black, Laurence Golding, Travis Hoppe, Charles Foster, Jason Phang, Horace He, Anish Thite, Noa Nabeshima, Shawn Presser, and Connor Leahy.
\newblock The pile: An 800gb dataset of diverse text for language modeling, 2020.

\bibitem[Gehman et~al.(2020)Gehman, Gururangan, Sap, Choi, and Smith]{gehman-etal-2020-realtoxicityprompts}
Samuel Gehman, Suchin Gururangan, Maarten Sap, Yejin Choi, and Noah~A. Smith.
\newblock {R}eal{T}oxicity{P}rompts: Evaluating neural toxic degeneration in language models.
\newblock In Trevor Cohn, Yulan He, and Yang Liu, editors, \emph{Findings of the Association for Computational Linguistics: EMNLP 2020}, pages 3356--3369, Online, November 2020. Association for Computational Linguistics.
\newblock \doi{10.18653/v1/2020.findings-emnlp.301}.
\newblock URL \url{https://aclanthology.org/2020.findings-emnlp.301}.

\bibitem[Geiger et~al.(2021)Geiger, Lu, Icard, and Potts]{geiger2021causal}
Atticus Geiger, Hanson Lu, Thomas Icard, and Christopher Potts.
\newblock Causal abstractions of neural networks, 2021.

\bibitem[Geva et~al.(2022)Geva, Caciularu, Wang, and Goldberg]{geva-etal-2022-transformer}
Mor Geva, Avi Caciularu, Kevin Wang, and Yoav Goldberg.
\newblock Transformer feed-forward layers build predictions by promoting concepts in the vocabulary space.
\newblock In Yoav Goldberg, Zornitsa Kozareva, and Yue Zhang, editors, \emph{Proceedings of the 2022 Conference on Empirical Methods in Natural Language Processing}, pages 30--45, Abu Dhabi, United Arab Emirates, December 2022. Association for Computational Linguistics.
\newblock \doi{10.18653/v1/2022.emnlp-main.3}.
\newblock URL \url{https://aclanthology.org/2022.emnlp-main.3}.

\bibitem[Havrilla et~al.(2023)Havrilla, Zhuravinskyi, Phung, Tiwari, Tow, Biderman, Anthony, and Castricato]{havrilla-etal-2023-trlx}
Alexander Havrilla, Maksym Zhuravinskyi, Duy Phung, Aman Tiwari, Jonathan Tow, Stella Biderman, Quentin Anthony, and Louis Castricato.
\newblock trl{X}: A framework for large scale reinforcement learning from human feedback.
\newblock In \emph{Proceedings of the 2023 Conference on Empirical Methods in Natural Language Processing}, pages 8578--8595, Singapore, December 2023. Association for Computational Linguistics.
\newblock \doi{10.18653/v1/2023.emnlp-main.530}.
\newblock URL \url{https://aclanthology.org/2023.emnlp-main.530}.

\bibitem[Kaufmann et~al.(2024)Kaufmann, Weng, Bengs, and Hüllermeier]{kaufmann2024survey}
Timo Kaufmann, Paul Weng, Viktor Bengs, and Eyke Hüllermeier.
\newblock A survey of reinforcement learning from human feedback, 2024.

\bibitem[Lee et~al.(2024)Lee, Bai, Pres, Wattenberg, Kummerfeld, and Mihalcea]{lee2024mechanistic}
Andrew Lee, Xiaoyan Bai, Itamar Pres, Martin Wattenberg, Jonathan~K. Kummerfeld, and Rada Mihalcea.
\newblock A mechanistic understanding of alignment algorithms: A case study on dpo and toxicity, 2024.

\bibitem[Li et~al.(2023)Li, Guo, Fan, Xu, Huang, Meng, and Song]{li2023multistep}
Haoran Li, Dadi Guo, Wei Fan, Mingshi Xu, Jie Huang, Fanpu Meng, and Yangqiu Song.
\newblock Multi-step jailbreaking privacy attacks on chatgpt, 2023.

\bibitem[Maas et~al.(2011{\natexlab{a}})Maas, Daly, Pham, Huang, Ng, and Potts]{imdb2011dataset}
Andrew~L. Maas, Raymond~E. Daly, Peter~T. Pham, Dan Huang, Andrew~Y. Ng, and Christopher Potts.
\newblock Learning word vectors for sentiment analysis.
\newblock In \emph{Proceedings of the 49th Annual Meeting of the Association for Computational Linguistics: Human Language Technologies}, pages 142--150, Portland, Oregon, USA, June 2011{\natexlab{a}}. Association for Computational Linguistics.
\newblock URL \url{http://www.aclweb.org/anthology/P11-1015}.

\bibitem[Maas et~al.(2011{\natexlab{b}})Maas, Daly, Pham, Huang, Ng, and Potts]{maas-etal-2011-learning}
Andrew~L. Maas, Raymond~E. Daly, Peter~T. Pham, Dan Huang, Andrew~Y. Ng, and Christopher Potts.
\newblock Learning word vectors for sentiment analysis.
\newblock In Dekang Lin, Yuji Matsumoto, and Rada Mihalcea, editors, \emph{Proceedings of the 49th Annual Meeting of the Association for Computational Linguistics: Human Language Technologies}, pages 142--150, Portland, Oregon, USA, June 2011{\natexlab{b}}. Association for Computational Linguistics.
\newblock URL \url{https://aclanthology.org/P11-1015}.

\bibitem[Ng et~al.(1999)Ng, Harada, and Russell]{Ng1999PolicyIU}
A.~Ng, Daishi Harada, and Stuart~J. Russell.
\newblock Policy invariance under reward transformations: Theory and application to reward shaping.
\newblock In \emph{International Conference on Machine Learning}, 1999.
\newblock URL \url{https://api.semanticscholar.org/CorpusID:5730166}.

\bibitem[nostalgebraist(2020)]{nostalgebraist_interpreting_gpt}
nostalgebraist.
\newblock Interpreting gpt: The logit lens, 2020.
\newblock Accessed on 2024-04-06.

\bibitem[Olah(2022)]{olah2022mechanistic}
Chris Olah.
\newblock Mechanistic interpretability, variables, and the importance of interpretable bases.
\newblock \url{https://www.transformer-circuits.pub/2022/mech-interp-essay}, 2022.

\bibitem[Rao et~al.(2024)Rao, Vashistha, Naik, Aditya, and Choudhury]{rao2024tricking}
Abhinav Rao, Sachin Vashistha, Atharva Naik, Somak Aditya, and Monojit Choudhury.
\newblock Tricking llms into disobedience: Formalizing, analyzing, and detecting jailbreaks, 2024.

\bibitem[Sanh et~al.(2020)Sanh, Debut, Chaumond, and Wolf]{sanh2020distilbert}
Victor Sanh, Lysandre Debut, Julien Chaumond, and Thomas Wolf.
\newblock Distilbert, a distilled version of bert: smaller, faster, cheaper and lighter, 2020.

\bibitem[Schulman et~al.(2017{\natexlab{a}})Schulman, Levine, Moritz, Jordan, and Abbeel]{schulman2017trust}
John Schulman, Sergey Levine, Philipp Moritz, Michael~I. Jordan, and Pieter Abbeel.
\newblock Trust region policy optimization, 2017{\natexlab{a}}.

\bibitem[Schulman et~al.(2017{\natexlab{b}})Schulman, Wolski, Dhariwal, Radford, and Klimov]{schulman2017proximal}
John Schulman, Filip Wolski, Prafulla Dhariwal, Alec Radford, and Oleg Klimov.
\newblock Proximal policy optimization algorithms, 2017{\natexlab{b}}.

\bibitem[Sheng et~al.(2019)Sheng, Chang, Natarajan, and Peng]{sheng-etal-2019-woman}
Emily Sheng, Kai-Wei Chang, Premkumar Natarajan, and Nanyun Peng.
\newblock The woman worked as a babysitter: On biases in language generation.
\newblock In Kentaro Inui, Jing Jiang, Vincent Ng, and Xiaojun Wan, editors, \emph{Proceedings of the 2019 Conference on Empirical Methods in Natural Language Processing and the 9th International Joint Conference on Natural Language Processing (EMNLP-IJCNLP)}, pages 3407--3412, Hong Kong, China, November 2019. Association for Computational Linguistics.
\newblock \doi{10.18653/v1/D19-1339}.
\newblock URL \url{https://aclanthology.org/D19-1339}.

\bibitem[Skalse et~al.(2022)Skalse, Howe, Krasheninnikov, and Krueger]{skalse2022hacking}
Joar Skalse, Nikolaus Howe, Dmitrii Krasheninnikov, and David Krueger.
\newblock Defining and characterizing reward gaming.
\newblock In S.~Koyejo, S.~Mohamed, A.~Agarwal, D.~Belgrave, K.~Cho, and A.~Oh, editors, \emph{Advances in Neural Information Processing Systems}, volume~35, pages 9460--9471. Curran Associates, Inc., 2022.
\newblock URL \url{https://proceedings.neurips.cc/paper_files/paper/2022/file/3d719fee332caa23d5038b8a90e81796-Paper-Conference.pdf}.

\bibitem[Sutton and Barto(2018)]{sutton2018reinforcement}
Richard~S Sutton and Andrew~G Barto.
\newblock \emph{Reinforcement Learning: An Introduction}.
\newblock The MIT Press, second edition, 2018.
\newblock ISBN 978-0-262-03924-6.

\bibitem[Vaswani et~al.(2023)Vaswani, Shazeer, Parmar, Uszkoreit, Jones, Gomez, Kaiser, and Polosukhin]{vaswani2023attention}
Ashish Vaswani, Noam Shazeer, Niki Parmar, Jakob Uszkoreit, Llion Jones, Aidan~N. Gomez, Lukasz Kaiser, and Illia Polosukhin.
\newblock Attention is all you need, 2023.

\bibitem[Wei et~al.(2022)Wei, Tay, Bommasani, Raffel, Zoph, Borgeaud, Yogatama, Bosma, Zhou, Metzler, Chi, Hashimoto, Vinyals, Liang, Dean, and Fedus]{wei2022emergent}
Jason Wei, Yi~Tay, Rishi Bommasani, Colin Raffel, Barret Zoph, Sebastian Borgeaud, Dani Yogatama, Maarten Bosma, Denny Zhou, Donald Metzler, Ed~H. Chi, Tatsunori Hashimoto, Oriol Vinyals, Percy Liang, Jeff Dean, and William Fedus.
\newblock Emergent abilities of large language models, 2022.

\bibitem[Zhu et~al.(2015)Zhu, Kiros, Zemel, Salakhutdinov, Urtasun, Torralba, and Fidler]{bookscorpus2015}
Yukun Zhu, Ryan Kiros, Rich Zemel, Ruslan Salakhutdinov, Raquel Urtasun, Antonio Torralba, and Sanja Fidler.
\newblock Aligning books and movies: Towards story-like visual explanations by watching movies and reading books.
\newblock In \emph{2015 IEEE International Conference on Computer Vision (ICCV)}, pages 19--27, 2015.
\newblock \doi{10.1109/ICCV.2015.11}.

\end{thebibliography}
\bibliographystyle{plainnat}

\newpage

\end{document}